\documentclass{article}

\usepackage[utf8]{inputenc} 
\usepackage[T1]{fontenc}    
\usepackage{hyperref}       
\usepackage{url}            
\usepackage{xurl}  
\usepackage{authblk}
\usepackage{graphicx}
\usepackage{hhline}
\usepackage[small]{caption}
\usepackage[margin=3cm]{geometry} 

\title{Prompt Selection Matters: Enhancing Text Annotations for Social Sciences with Large Language
Models}

\author[1]{Louis Abraham\textsuperscript{*}}
\author[2]{Charles Arnal\textsuperscript{*}}
\author[3]{Antoine Marie\textsuperscript{*}}

\affil[1]{Université Paris 1 Panthéon-Sorbonne,
\href{https://scholar.google.fr/citations?user=c1Norb8AAAAJ&hl=en}{Google Scholar}}
\affil[2]{Universit{\'e} Paris-Saclay, CNRS, Inria,
\href{https://scholar.google.com/citations?user=Pre7QicAAAAJ&hl=en}{Google Scholar}}
\affil[3]{Institut Jean Nicod, Ecole Normale Sup{\'e}rieure, PSL-EHESS-CNRS, \href{https://scholar.google.com/citations?hl=en&user=immG3K4AAAAJ&view_op=list_works&gmla=AC6lMd8GlGzRqp_6G-IUKuSTzrB3fqoBZ7GYA1LXSinBhZpro5hyEevDZrJJLTqz9Sw0am1yjkHFdqU52HFNpoSA}{Google Scholar}}

\begin{document}

\maketitle

\def\thefootnote{*}\footnotetext{Equal contribution.}\def\thefootnote{\arabic{footnote}}

\begin{abstract}
Large Language Models have recently been applied to text annotation tasks from social sciences, equating or surpassing the performance of human workers at a fraction of the cost.
However, no inquiry has yet been made of the impact of prompt selection on labelling accuracy.
In this study, we show that performance greatly varies between prompts, and we apply the method of automatic prompt optimization to systematically craft high quality prompts.
We also provide the community with a simple, browser-based implementation of the method at \newline 
\url{https://prompt-ultra.github.io/}.

\end{abstract}

\section{Introduction}

Throughout the social sciences, many research questions are answered using annotation and classification of large volumes of text, such as tweets or Facebook comments. Researchers may be interested in knowing, for instance, how politically slanted (liberal vs. conservative) a claim or headline is; how emotional or hostile its tone is, or which one of the basic emotions it reflects (sadness, joy, anger, etc) \cite{barrett2016handbook, howpartisanispress2012, Strapparava2010, rasmussen2024super}.
Text annotation so far had to be performed by humans--experts, research assistants, or unskilled crowd workers who are likely to mislabel the data \cite{Li2024comparative}--or using machine learning models trained via supervised learning specifically for the task at hand. As a result, it was typically costly and time-consuming.

The recent progress of Large Language Models (LLMs) has opened new avenues and could revolutionize text mining by allowing huge volumes of text data to be analyzed in an unsupervised way in a matter of minutes  \cite{heseltine2024large, gilardi2023chatgpt, tornberg2023chatgpt, EvalAllYouNeed, DoanParty}. Early results suggest that LLMs can perform extremely well, extremely fast and at a reasonable cost, with levels of accuracy rivaling those of experts and superior to those of unskilled workers. 
Moreover, these results are achieved by off-the-shelf, general purpose models, that do not need any specialized training, unlike some pre-existing machine learning-based techniques.

Nonetheless, crucial aspects of the automatic annotation of text data using LLMs have not been studied yet.
In particular, few studies have considered the importance of prompt choice in this context: simple hand-crafted prompts, such as ``Does the following message express liberal or conservative views?'' or ``Is this tweet pro-life or pro-choice?'', are frequently used to annotate corpora.
However, it has been observed outside the context of text annotation that applying distinct yet similar versions of a given prompt to certain tasks can result in large differences in accuracy\footnote{Similar differences can be observed when using other performance metrics, such as F-scores.}, of the order of more than $10\%$ \cite{kojima2022large, few_shot_learners}; similar findings within the context of text annotation are reported in \cite{EvalAllYouNeed}.
Going from $10\%$ to $20\%$ of mislabelled data can greatly impact the quality of a study's conclusions, especially if the classification errors are biased (e.g., if all mislabelled tweets express conservative views).

Our main objectives in this paper are threefold. First, to help social better social scientists understand automatic text annotation using LLMs by providing a clear illustration of how it works. Second, to demonstrate the importance of prompt selection on performance, and to raise awareness about the issue of performance variability. Third, to explain to social scientists how to apply state-of-the-art prompt optimization methods used in the wider LLM community \cite{optimizer1, optimizer2} to their own classification tasks.

More precisely, our contributions are as follows:
\begin{itemize}
    \item We give a short, didactic overview of automatic text annotation for social sciences using LLMs; in particular, we draw attention to some shortcomings of the method that have not yet been discussed.
    \item We describe the principle of \textit{automatic prompt optimization}, as well as a simple implementation of the method.
    \item To the social science community, we provide a \textbf{website} that enables researchers to efficiently label their own datasets using LLMs using either their own hand-crafted prompts or a prompt optimization algorithm. This browser-based service can be found at  ~\url{https://prompt-ultra.github.io/}.
    \item We investigate and quantify the impact of prompt selection on performance across a range of standard classification tasks in social sciences using LLMs. To that end, we compare both hand-crafted prompts and automatically optimized prompts.
   \item We conclude that apparently similar prompts in natural language yield greatly varied performance levels. We also observe that automatic prompt optimization yields consistently good performance and beats prompt-crafting heuristics (such as using Chain of Thoughts prompts, see below) on most tasks.

\end{itemize}
All of the code used to perform the experiments will be made available on GitHub.

\section{Automatic text labelling using LLMs}\label{sec:general_presentation}

In this section, we present the recently proposed method of automatic text labelling using LLMs, and discuss some of its shortcomings.

State-of-the-art general purpose language models, such as GPT-4 \cite{GPT4}, have now achieved human-like performance on a variety of tasks. In particular, it has been recently demonstrated that they can be applied off-the-shelf, i.e., without needing any specialized training, to a variety of text-labelling tasks needed in social sciences.
Among other examples, they have been shown to reach around $95\%$ of accuracy\footnote{It is useful to note in passing that while it is desirable to maximize accuracy at classification tasks, it is generally impossible to reach 100\% in virtue of the fact that some statements are intrinsically ambiguous and thus impossible to classify even by professional raters.} on negativity detection in tweets and news articles  \cite{heseltine2024large}, around $75\%$ of accuracy on stance detection in tweets (largely outperforming untrained crowd workers) \cite{gilardi2023chatgpt}, and more than $92\%$ of accuracy on political orientation in tweets \cite{tornberg2023chatgpt}.

Up until one or two years ago, similar performances were either unachievable, or only achievable by models specifically trained on the task at hand, the creation of which was complex and time-consuming \cite{MLold1, MLold2, MLold3, pennacchiotti2011machine, devatine:hal-04249724}.

Let us give a quick explanation of how to automatically label text using LLMs. Though variants are possible, the simplest method is to simply ask in simple and clear terms an LLM-based chatbot, such as ChatGPT \cite{ChatGPT}, to give a label to a given piece of text.
For instance:
\begin{verbatim}
[Social scientist]
Does the following tweet express liberal or conservative opinions? 
Output only "liberal" or "conservative" without quotes.

Please support our Capitol Police and Law Enforcement.
They are truly on the side of our Country. Stay peaceful!

[ChatGPT] conservative
\end{verbatim}

Typing each prompt by hand in the browser interface would be tedious. Instead, researchers would typically send such prompts through the API using e.g., a simple Python script\footnote{Such a script can be found in our code base.}, allowing thousands of statements to be classified in a matter of minutes.
Each message would have to be processed in a fresh instance of the client chat, to avoid a progressive biasing of the model by the conversation history. 
A small percentage of answers might not be correctly formatted (the LLM might respond ``The message is conservative'' or ``Conservative.'' or ``right-wing'' instead of ``conservative''), in which case one would simply have to ask the chatbot again.
To get an estimate of the accuracy of the labels thus produced, the researcher should ideally have a small subset of the dataset be annotated by experts (typically a few hundreds to a few thousand items of text), so that the labels provided by the LLM can be compared to those provided by experts, which serve as ground truth for the task. The degree of agreement between the two sets of labels can then serve as an estimate of the accuracy of the LLM's annotations.

The perks of using LLMs to classify text, compared to manual annotation by crowd workers or experts, are immense. It is easy to implement, fast and comparatively inexpensive (thousands of short or medium length texts can be labelled in a few minutes for less than 10\$). It is also significantly less time-consuming and easier to set up compared to earlier machine learning-based methods.

\subsection{Limitations of automatic text labelling}

As for any method, automatic text annotation using LLMs suffers from some limitations, however, which have not been extensively discussed in the literature.
Though early studies are very encouraging, the range of tasks that have been studied so far remains limited; the vast majority of them consists in attributing one of a handful of labels to short- to medium-length texts (typically social media posts or news headlines).
Moreover, early results find that accuracy tends to be lower in languages other than English \cite{heseltine2024large}, which is coherent with general observations regarding LLMs' performances.
LLMs are also known to suffer from potential biases, including political ones \cite{bias1,bias2,bias3}; those could result in systematic errors on certain tasks that would have a greater impact on the research performed on the annotated dataset than the raw accuracy might suggest. As an example, consider an LLM tasked with labelling a set of tweets as Republican-leaning or Democrat-leaning ; if all mislabelled tweets are Republican tweets erroneously labeled as Democrat, the conclusions drawn from the dataset might be wrong, despite the overall accuracy being high.

The issue can also be exacerbated by conscious moderating efforts by the LLM's creators: various techniques are in place to ensure that the answers provided by LLM-based chat agents avoid using stereotypes seen as offensive \cite{Achiam2023GPT4TR}, which can negatively affect their performance on politically sensitive topics like race.



Other problems arise not from the intrinsic nature of LLMs, but from the way in which they are currently being trained and deployed.
State-of-the-art LLMs, such as GPT-4, are trained on immense datasets of text scrapped from the web, and regularly updated in a similar fashion.
 Beyond the possible privacy issues, this has crucial implications regarding the performance of LLMs: they will often perform better on data on which they have been trained than on yet unseen data (e.g., data produced after their training) \cite{MLIntro}.
 As a result, the same LLM might be better at classifying a certain set of tweets than a more recent set it has not yet ``seen''.
 Since the exact nature of the training data is typically not shared with the general public, such phenomena are very hard to predict - see some of our observations in Section \ref{sec:results}. In particular, we suspect that some of the most impressive accuracy scores achieved by LLMs on annotation tasks (and reported in the academic literature) might be partially explained by this.
 This problem, in conjunction with the recurrent updating of models, creates important issues of replicability: results might vary greatly between apparently similar tasks, or when applying different versions of the same model to the same task. Note that this issue can be partially offset by using ``frozen'' LLMs, i.e., by keeping a copy of a certain version of an LLM at a certain point in time and using it on all tasks.


\subsection{Related Works on Prompting for Text Classification}

Since the 2000s, sociologists and political scientists have applied machine learning techniques, such as Support Vector Machines, Naive Bayes classifiers, or Maximum Entropy models, to the classification of large volumes of text (e.g. \cite{MLold1, MLold2, MLold3, pennacchiotti2011machine, devatine:hal-04249724, WilkersonComputerizedtextanalysis, MLIntro}).
However, these methods face several limitations \cite{QianFromTraditionaltoDeepLearning}, such as struggling with understanding context and informal language patterns common in social media posts, and their performance heavily depends on curated training datasets and domain-specific feature engineering.

A few years ago, Large language models (LLMs) have become available to social scientists and they are greatly facilitating text classification tasks in social sciences \cite{heseltine2024large, gilardi2023chatgpt, tornberg2023chatgpt, EvalAllYouNeed, Political-LLM-multipleauthors} (but see \cite{OllionMindllmhype} for a more critical perspective).
Early results show that off-the-shelves LLMs can perform extremely well, with accuracy levels sometimes equating those of human experts, while being comparatively cheap and easy to deploy.
In political discourse analysis in particular, they have been shown to successfully identify political slant, implicit biases, dog whistles, or rhetorical devices that simpler keyword-based approaches would miss \cite{DoanParty, heseltine2024large, gilardi2023chatgpt, tornberg2023chatgpt, EvalAllYouNeed, Political-LLM-multipleauthors}. 

The most straightforward way to annotate using LLMs is to ask them the same question, or \textit{prompt}, as would be asked to human coders
\cite{gilardi2023chatgpt, Anglin2024, he-etal-2024-annollm, heseltine2024large}.
However, this approach suffers from the models' sensitivity to the precise wording of the instructions: performances have been observed to improve in response to various ``tricks'' such as providing the LLM with additional context for the task, giving it examples of correct answers, and asking the LLM to explain its response or to answer while pretending to be a political analyst (see Subsection~\ref{subsec:handcrafted_prompts}).
Since then, a literature has emerged documenting such prompt engineering techniques (see e.g.,
\cite{few_shot_learners, Han2021PTRPT, kojima2022large, Lampinen2022CanLM, Wei2022ChainOT, Battle2024TheUE}, and the surveys \cite{survey1,survey2}).
However, novel prompts tend to be explored in an unsystematic manner and there is no guarantee that a given technique will perform well in an untested setting. This issue can be partially circumvented by annotating part of the data through other means (e.g., human coders) and using this data sample to evaluate the performance of various prompts, before applying the best prompt to the remainder of the data \cite{Anglin2024, EvalAllYouNeed}. Doing this is time-consuming, however. It also requires ground truth labels for evaluating prompt performance (which can be costly to obtain), and relies on the researcher’s ability to intuit which candidate formulations would make for better prompts.

As an answer to this problem, a more principled approach has recently been put forward, called Automatic prompt optimization or Automatic prompt engineering, which consists in using LLMs themselves to automatically engineer good prompts \cite{optimizer1, optimizer2,shum-etal-2023-automatic, APEER, RePrompt}. We employ and describe such a procedure below, in Subsection \ref{subsec:automatic_prompt_optimization}.


\section{Prompt Ultra, our automatic dataset labelling app}

Though using LLMs to label datasets demands significantly fewer technical skills compared to earlier machine learning-based methods, a basic understanding of coding is still necessary.
We are providing a browser app, \textbf{Prompt Ultra}, which can be used to automatically label datasets without any coding being required. It is freely available at \url{https://prompt-ultra.github.io/}.

Briefly summarized, one can use the app (and a ChatGPT access key, easily purchasable on their website \url{https://help.openai.com/en/}) to label any textual dataset with a prompt of one's choice. One can also evaluate and compare the performance of various prompts, as well as apply the automatic prompt optimization algorithm from Subsection \ref{subsec:automatic_prompt_optimization}.
More complete descriptions of the app are available in the Appendix, as well as on the website itself.

\begin{figure}[h]
    \centering
    \includegraphics[width=0.6\textwidth]{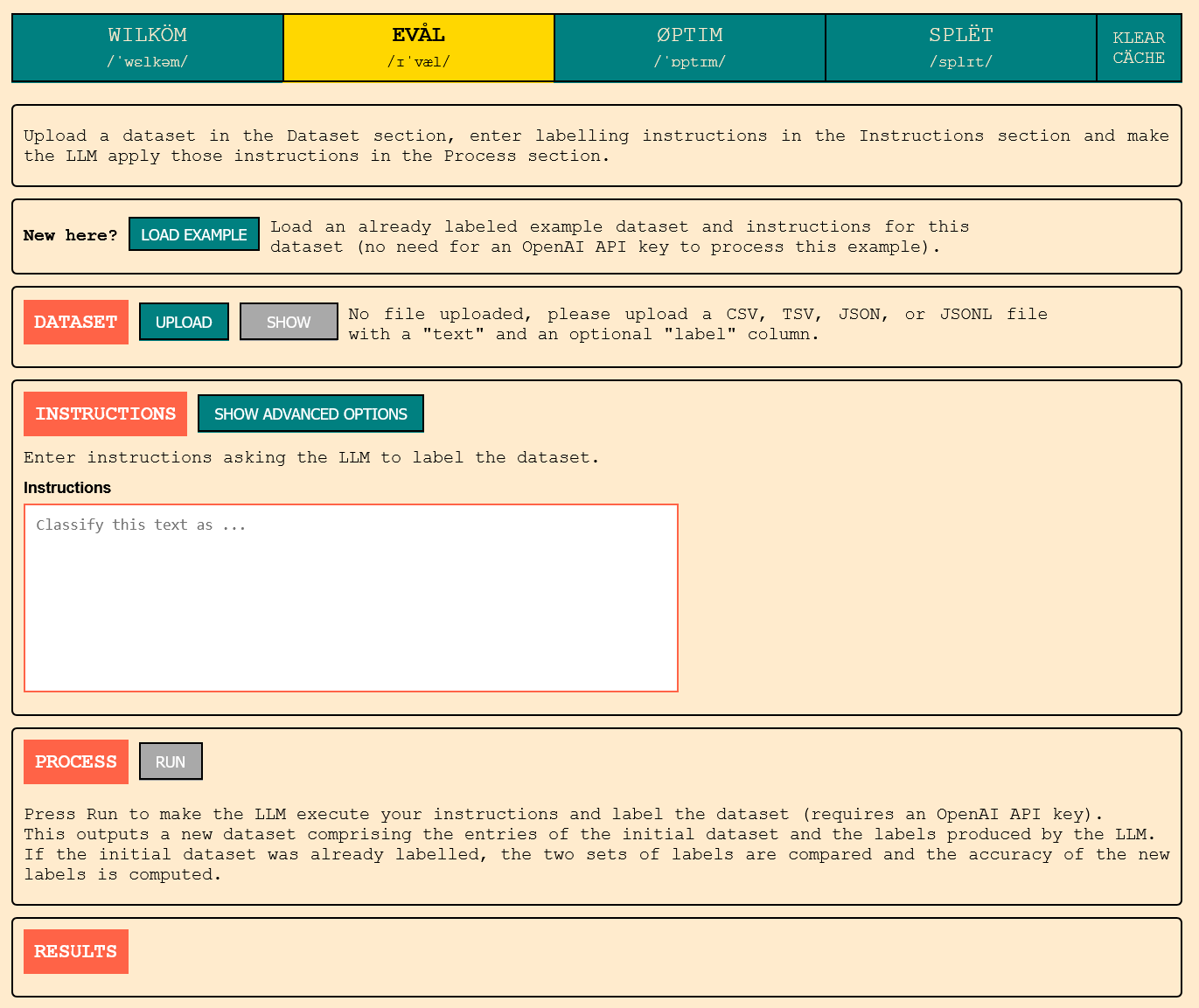}
    \caption{A screenshot of one of the website's tabs.}
    \label{fig:Eval_main_text}
\end{figure}

\section{Experiments}

We evaluate various hand-crafted prompts and automatically optimized prompts (see Subsections \ref{subsec:handcrafted_prompts} and \ref{subsec:automatic_prompt_optimization} below) on a range of datasets and classical social sciences tasks using OpenAI's GPT-3.5 Turbo's API \cite{gpt35} (see our code base for details).
Our objective is twofold: first, we want to measure the impact of the precise formulation of the prompt on GPT's accuracy by comparing  performance across various prompts. The goal here is not to find some optimal prompt crafting technique that would systematically outperform all others (such panacea is unlikely to exist, as suggested by our experimental results further below); in particular, we do not claim to have examined every possible prompt-crafting trick. Rather, we only want to see whether two reasonable, semantically similar prompts can result in significantly different accuracies on a given task.
Second, we want to check whether automatic prompt optimization can help achieve consistently good (though not necessarily optimal) results on all tasks without the need for manual tweaking by the experimenter.

\subsection{Datasets and tasks}\label{subsec:datasets}
We have selected a range of diverse yet typical annotation tasks and datasets.
When the sets have a predefined train and test sets split, we only use the test set for reasons explained in Section \ref{sec:results}.

\textbf{TweetEval - hate, emotion, sentiment, offensive (TE-hate, TE-emotion, TE-sent, TE-off)}

\url{https://github.com/cardiffnlp/tweeteval}

TweetEval \cite{tweeteval_all} consists of seven heterogeneous tasks performed on Twitter data in English, of which we have selected four:
\begin{itemize}
    \item Hate detection (hateful or non-hateful); the set contains $2'970$ tweets \cite{tweeteval_hate}.
    \item Emotion recognition (anger, joy, optimism or sadness); the set contains $1'421$ tweets \cite{tweeteval_emotion}.
    \item Sentiment recognition (negative, neutral, positive); the set contains $12'284$ tweets, and we randomly draw and use $10'000$ of them \cite{tweeteval_sentiment}.
    \item Offensive language detection (non-offensive, offensive); the set contains $ 860$ tweets \cite{tweetevaloffensive}.
\end{itemize}

\textbf{Tweet Sentiment Multilingual (TML-sent)}

\url{https://github.com/cardiffnlp/xlm-t/tree/main/data/sentiment}

Tweet Sentiment Multilingual \cite{tweetsentimentmultilingual} is a dataset of tweets in $8$ different languages (Arabic, English, French, German, Hindi, Italian, Portuguese, Spanish) with sentiment analysis labels (negative, neutral and positive). The test set contains $6'960$ messages.

\textbf{Article Bias Prediction (AS-pol)}

\url{https://github.com/ramybaly/Article-Bias-Prediction}

A dataset of $37554$ news articles from major US newspapers \cite{allsides_dataset}, with labels representing their political inclination (left, center\footnote{Note that the centrist label describes articles that are biased towards a centrist political ideology, and not articles that lack political bias.} or right), from which we randomly sample $10'000$ articles.

\textbf{Liberals vs Conservatives on Reddit (LibCon)}

\url{https://www.kaggle.com/datasets/neelgajare/liberals-vs-conservatives-on-reddit-13000-posts}

A dataset of $13'000$ Reddit posts in English collected from liberal- and conservative-leaning subreddits \cite{libcon}. We randomly select and use $10'000$ of them.

\subsection{Hand-crafted prompts}\label{subsec:handcrafted_prompts}

As noted in our review of related works, it has been observed that various tricks can help increase LLMs' performance. As an example, simply adding ``Let's think step by step'' at the end of a prompt can result in impressive improvements \cite{kojima2022large}.
Though there is no systematic theory of ``good prompting'', many such  tricks have been catalogued (see e.g., \cite{prompt_engineering_guide}).
We have applied several of these standard techniques to craft five different prompts for each task; below are examples from the \textbf{TweetEval - hate} task.
The other prompts can be found in the accompanying code.

\begin{itemize}
    \item \textbf{Simple} - A simple, minimalist prompt:

    \textit{Classify the following message as hateful if it contains hate speech or non-hateful if it does not contain hate speech. Output only “hateful” or “non-hateful” without quotes.}
    \item \textbf{Explanations} - A prompt enriched with additional explanations (see e.g. \cite{Lampinen2022CanLM}):

    \textit{You will be given a tweet. Your task is to use your knowledge of sociology, politics and culture to make an educated guess on whether the tweet contains hate speech. Hate speech includes any speech that expresses hate or encourages violence towards a person or group based on something such as race, religion, sex, sexual orientation or socio-economic situation. For example, messages that aggressively and unfairly target women or immigrants constitute hate speech. If the message does not contain enough information to correctly classify it, just make your best guess. If the message contains hate speech, output “hateful” without quotes. If the message does not contain hate speech, output “non-hateful” without quotes.
Output only “hateful” or “non-hateful” without quotes.}

\item \textbf{Examples} - A prompt containing the same explanations, as well as some examples of correctly classified messages (see e.g. \cite{few_shot_learners}).

\textit{You will be given a tweet. Your task is to [...]}

\textit{Output only “hateful” or “non-hateful” without quotes.}

\textit{Here are four examples of tweets with the correct classification :}

\textit{Women are equal and deserve respect. Just kidding, they should suck my dick.}

\textit{hateful}

\textit{I've been called a son of a whore and a cunt today already. It's not even 1 yet}

\textit{non-hateful}

\textit{[\ldots]}

\item \textbf{Roleplay} - A prompt containing the same explanations and asking the chatbot to answer the question while roleplaying as a political analyst (see e.g. \cite{Battle2024TheUE}).

\textit{You will be given a tweet. Pretend that you are a political analyst and decide whether the tweet contains hate speech. Hate speech includes [\ldots]}

\item \textbf{Chain of thoughts (CoT)} - A prompt with the same explanations that asks the LLM to first explain its reasoning, then answer the question (see \cite{Wei2022ChainOT, kojima2022large}). The very last word of the output is used as a label.

\textit{
You will be given a tweet. Your task is to [\ldots] constitute hate speech.
First explain your reasoning. Then start a new line, and output “hateful” without quotes if the message contains hate speech, and output “non-hateful” without quotes if the message does not contain hate speech.
The only word on the new line must be either “hateful” without quotes or “non-hateful” without quotes.}
\end{itemize}

\subsection{Automatic prompt optimization (APO)}\label{subsec:automatic_prompt_optimization}

A more systematic way to craft good prompts has been recently proposed: automatic prompt optimization \cite{optimizer1, optimizer2,shum-etal-2023-automatic} (also called automatic prompt engineering).

The core idea is to ask an LLM to repeatedly rephrase prompts, then to select the one that offers the best performance.
We translated this general concept into our specific text labelling setting as follows:
we first require a subset of the dataset to have already been labelled by human annotators; typically, a few hundred or a few thousand messages.
We then start from any reasonable prompt for the task, e.g., ``Classify the following message as hateful if it contains hate speech or non-hateful if it does not contain hate speech.'' in the case of the hate detection task.
We ask the LLM to reformulate the prompt several times, e.g. with ``Generate a variation of the following instruction while keeping the semantic meaning.'' 
We then repeat the following steps as many times as desired: we evaluate the current set of prompts on the labelled subset, we keep only the top prompts (in terms of scores), and we ask the LLM to reformulate them to create a new generation of prompts (in addition with the top prompts themselves). The best prompt of the last generation is then used to label the remainder of the dataset. Before that, one can estimate its associated accuracy by testing it on another pre-labelled subset, distinct from the one used during the optimization process to avoid any bias caused by a lack of independence.

In our experiments and for a given task, each generation had $8$ prompts; each was evaluated on a fixed subset of $400$ samples from the dataset. The top $2 $ prompts were kept, and each was reformulated $3$ times to generate a new generation of $2+2\times 3 = 8$ prompts. We repeated this process for $15$ generations, after which we kept the best performing prompt of the last generation.
Finally, we evaluated this prompt on the remainder of the dataset (without the $400$ prompts, to guarantee an unbiased estimate of the prompt's accuracy). 

The method is both conceptually simple, and easy to implement; its main a priori downside compared to using hand-crafted prompts is the increased number of calls to the chatbot API required, though the costs remain almost negligible.  It can be tested on our free browser-based service, which we further describe in Appendix \ref{app:online_app}.

\section{Results and discussion}
\label{sec:results}

\begin{figure}[h!]
    \centering
    \includegraphics[width=\linewidth]{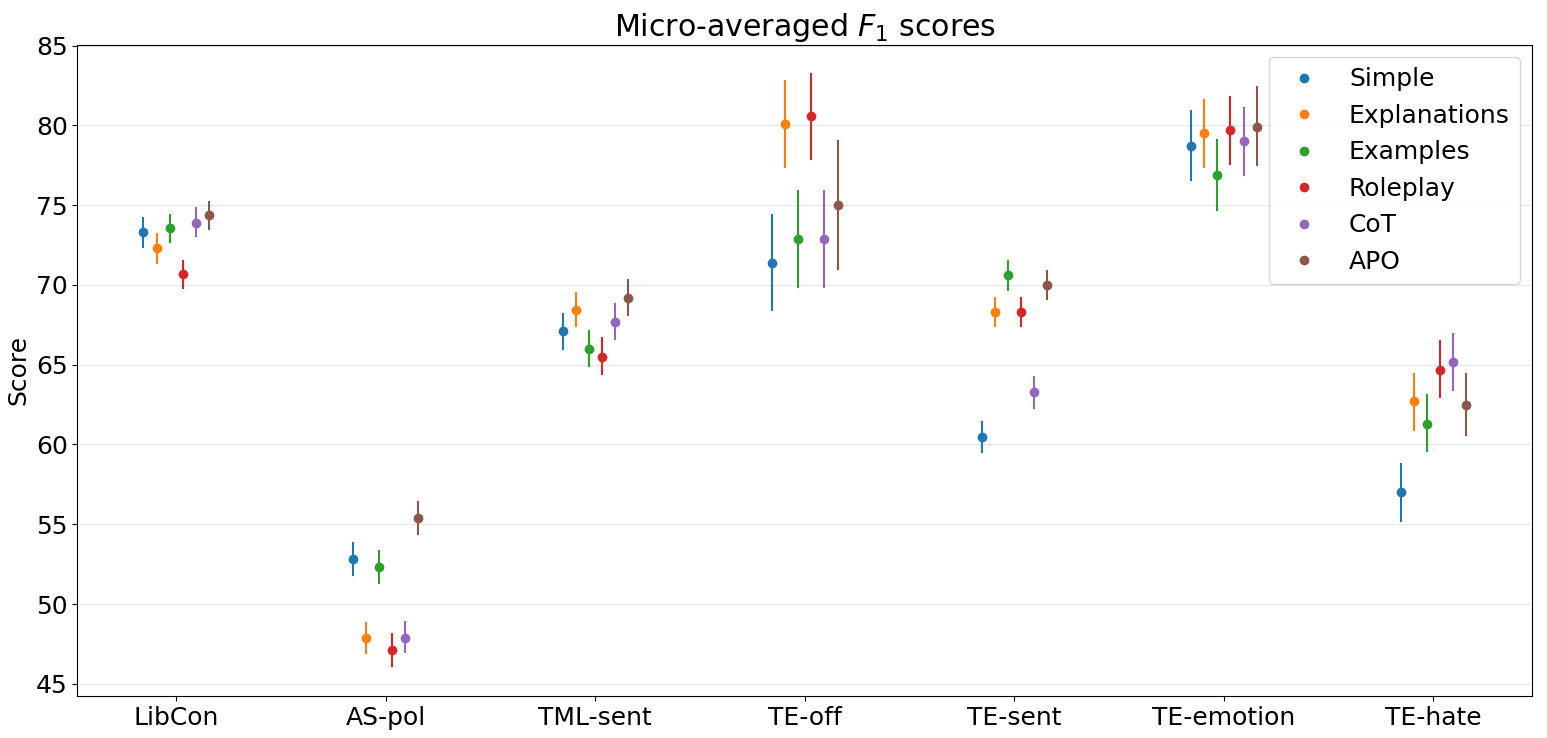} %
    \caption{Micro-averaged $F_1$ scores  (in $\%$)  of the hand-crafted prompts and of the best prompt
	obtained using automatic prompt optimization (APO) on each of the datasets and
	tasks described in Subsection \ref{subsec:datasets}. $95\%$~confidence intervals are represented. The same results are reported as a numerical table in the Appendix.}
    \label{fig:main_results}
\end{figure}

\textbf{Inter-prompts performance variability}

We report in Figure \ref{fig:main_results} the micro-averaged $F_1$ scores 
on each task of our five types of prompts and of the best prompt generated using automatic prompt generation, as well as their estimated 95\% confidence intervals \cite{ConfidenceIntervals, jacobgildenblatconfidenceinterval}.
We observe that the choice of prompt has a significant impact on performance for most tasks: depending on the task, the best prompt results in an observed $F_1$ score that is up to $18\%$ higher than that of the worst prompt.
This confirms our claim that careful prompt selection is crucial.

We also note that there is no "miracle" hand-crafted prompt that consistently outperforms all the others. Some heuristics and specific wordings work well on certain tasks, and poorly on others.
This is in line with observations made in the literature; as an example, adding ``Let’s think step by step'' at the end of a prompt is shown to outperform adding the semantically similar ``Let’s work this out in a step by step way to be sure we have the right answer'' for the task studied in \cite{optimizer2} (see Table 1 thereof), while the converse is true in \cite{optimizer1} (Table 7).

\textbf{Automatic prompt optimization outperforms hand-crafted prompts}

We now turn to the specific performances of Automatic prompt optimization (APO). APO beats hand-crafted prompts at four tasks; among the remaining three, it is second best for TE-sent (with a score nearly equal to that of the best hand-crafted prompt), and is roughly at the level of the median of the hand-crafted prompts for TE-off and TE-hate.
This shows that automatically optimizing prompts results in consistently good performance without the need for prior testing of various hand-crafted prompts on a subset of the dataset of interest.
Note that the expected performance of APO could be improved by increasing the number of samples on which each generation is tested (the $400$ samples mentioned in Subsection \ref{subsec:automatic_prompt_optimization}), in exchange for an increase in computational costs.

Interestingly, and in line with the remarks of the previous paragraph, the final prompts generated using APO do not differ much from the ones used to start the optimization process, despite the often large gap between their associated accuracies\footnote{We report the best prompt returned by the optimization process for each task in the Appendix.}.
For TE-hate, for example, the initial prompt is 
``Classify the following message as hateful if it contains hate speech or non-hateful if it does not contain hate speech. Output only “hateful” or “non-hateful” without quotes'' and the automatically optimized prompt is ``Check for hate speech in the following message to determine if it is hateful, then classify it as either "hateful" or "non-hateful"''.

\textbf{Is ChatGPT cheating?}

\begin{figure}[h!]
    \centering
    \includegraphics[width=\linewidth]{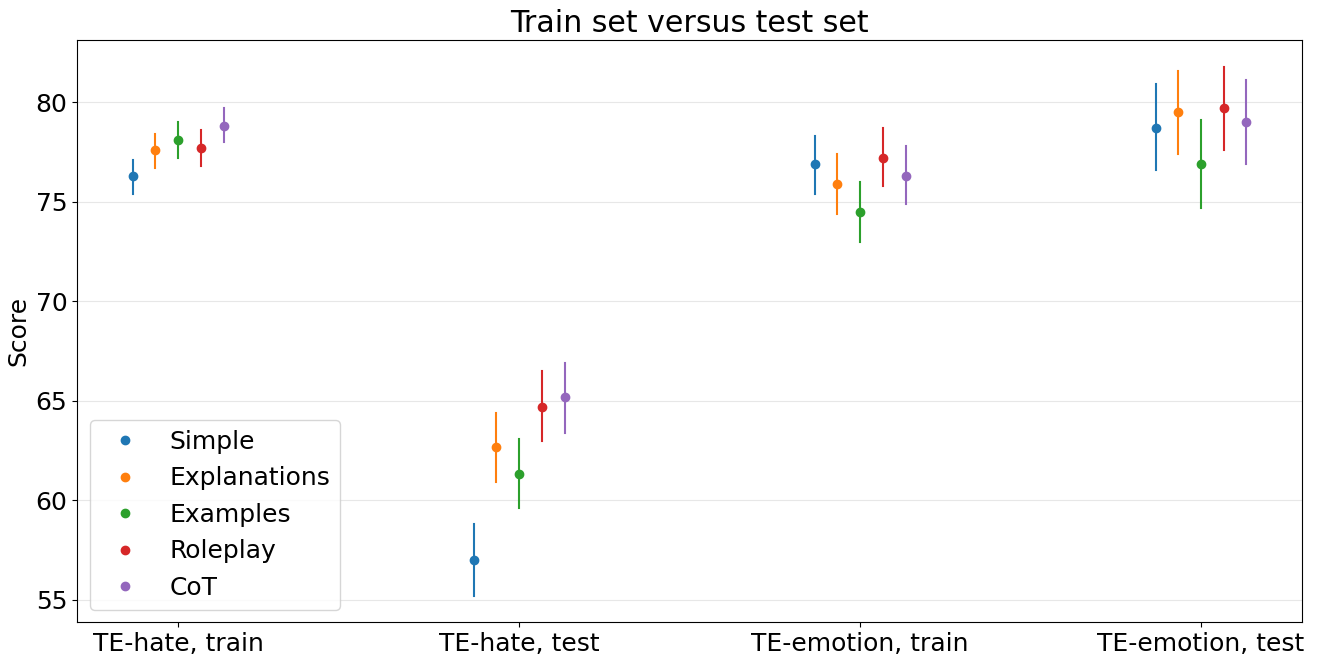} %
    \caption{Micro-averaged $F_1$ scores (in $\%$) of the hand-crafted prompts
    on the train set and on the test set of the TE-hate and TE-emotion datasets. $95\%$  confidence intervals are represented. The same results are reported as a numerical table in the Appendix.}
    \label{fig:test_vs_train}
\end{figure}

We have observed a curious phenomenon: as shown in Figure \ref{fig:test_vs_train}, all prompts result in much higher $F_1$ score when tested on the training set of the TE-hate dataset than on its test set (where ``training'' and ``test'' refer to the split made by the creators of the dataset \cite{tweeteval_hate}).
This is a priori surprising, as both subsets should be drawn from the same distribution, and as such our prompts should not perform any better on the training set than on the test set\footnote{Note that there is no issue of overfitting here, as the prompts were not ``trained'' in any way on what we call the training set in this context.}.
Our explanation is that the training set of TE-hate is part of the enormous amount of data on which GPT-3.5 Turbo, the model which we used, has been trained; as a result, it performs better on it than on the test set (which was either not included, or included in a different fashion). As the exact data on which the model was trained was not made public, we cannot confirm this hypothesis with absolute certainty.
As a result, we can say that ChatGPT is ``cheating'' when labelling the training set of TE-hate, in the sense that its performance is superior to what it would be on a similar dataset to which it did not have access during its training (e.g. data that was produced after its training). 
This explains why we used the test splits of our datasets whenever possible, in the hope that as (we assume it to be the case) for TE-hate, they have not been included in the training set of the model, making for a fairer assessment of the capabilities of the method.

This problem makes it hard to predict future performances based on past experiments, as we have no guarantee as to what was and was not used in the models' training. Note for example that a similar gap in performance between the training set and the test set is not observable for TE-emotion.
Ideally, one should only test the models on data that was produced after their training. This could however prove difficult and overly restrictive: curating useful datasets is a challenging and time-consuming task in itself, and most datasets available online are at least a few years old, thereby predating all last generation LLMs.

As a side note, this phenomenon could explain why we have observed slightly weaker performances on average than those reported in recent literature \cite{heseltine2024large, gilardi2023chatgpt, tornberg2023chatgpt} on similar tasks: we suspect that the authors of these articles might not have been aware of this pitfall, and failed to take measures to circumvent it. 
It could also be due to intrinsic differences in difficulty between the datasets used, as we tried to include both easy and hard datasets for diversity (in particular, it has been noted before, e.g., in \cite{heseltine2024large}, that performance drops for very long texts, such as the articles from the AS-pol dataset).
We cannot verify this hypothesis, as (most of) the datasets used in the references that we cite are not available anymore due to a policy change from the social media platform X.

\textbf{A note on our choice of datasets and task complexity}

We selected on purpose tasks that are commonly performed by social scientists and natural language processing researchers so as to make the relevance of our claims (variation in prompt performance and the benefits of automatic optimization) immediately apparent to both communities. A downside of this choice is that though distinct, the tasks considered are somewhat similar and inevitably simple. Another limitation of our datasets is that though some tasks are harder than others, none are hard enough to present a real challenge to a well-informed human being. In future work, it would be interesting to measure the accuracy and sensitivity of LLMs to the prompts chosen both on a greater variety of tasks and in limit cases where even human experts might struggle.
As a final remark, note that due to the very nature of the tasks at hand, some labels in our datasets are not unambiguously correct: it is for instance easy to find tweets in the TE-emotion dataset that could be classified as either joyful or optimistic.
This fact, while not unique to our datasets, necessarily has an impact on the performances reported.

\section{Conclusion}

This paper meant to illustrate the importance of prompt selection and the effectiveness of automatic prompt optimization within the context of text annotation tasks frequently used in the social sciences.
In particular, our results suggest that the prompt optimization procedure described in Subsection \ref{subsec:automatic_prompt_optimization} can be applied off-the-shelf to various tasks to achieve accuracies that equal or exceed those obtained using hand-crafted prompts.
This prompt optimization method is made particularly easy by our user-friendly implementation\footnote{In particular, it allows for the replication of all of our experiments without the need for any coding.}, which is freely available at 
\url{https://prompt-ultra.github.io/}.

Nonetheless, we have also seen that any automatic annotation procedure relying on last generation LLMs raises important questions of replicability due to their training processes (not to mention environmental and privacy-related concerns).

Among other potential future research directions, it would be particularly interesting to test whether LLMs tasked with labelling a corpus can give robust justifications for their choices -- e.g. explain that they classified a given tweet as left-wing due to the presence of such and such opinions that are typically associated with the Left.
Though more sophisticated solutions are conceivable (e.g. using the LLMs' attention mechanisms, see \cite{attention}), directly asking the chatbot "Can you justify your decision?" would already make for an interesting experiment.
Getting LLMs to associate a (reliable) confidence score to its labels, so that human annotators can review the small percentage of labels the model was unsure of and increase accuracy, would also be of great interest.

\section{Data availability statement}

All data used in our analyses are linked in the "3.1 Datasets and tasks" section of the manuscript.

\section{Conflict of interest}

Nothing declared.

\bibliographystyle{alpha}
\bibliography{bibliography}

\newcommand{\etalchar}[1]{$^{#1}$}
\begin{thebibliography}{BDSMGN20}

\bibitem[AAS{\etalchar{+}}20]{MLold3}
Mohd~Zeeshan Ansari, M.B. Aziz, M.O. Siddiqui, H.~Mehra, and K.P. Singh.
\newblock Analysis of political sentiment orientations on twitter.
\newblock {\em Procedia Computer Science}, 167:1821--1828, 2020.
\newblock International Conference on Computational Intelligence and Data Science.

\bibitem[Alp10]{MLIntro}
Ethem Alpaydin.
\newblock {\em Introduction to Machine Learning}.
\newblock The MIT Press, 2nd edition, 2010.

\bibitem[AV24]{Anglin2024}
Kylie~L. Anglin and Claudia Ventura.
\newblock Automatic text classification with large language models: A review of openai for zero- and few-shot classification.
\newblock {\em Journal of Educational and Behavioral Statistics}, 2024.

\bibitem[BBF{\etalchar{+}}19]{tweeteval_hate}
Valerio Basile, Cristina Bosco, Elisabetta Fersini, Debora Nozza, Viviana Patti, Francisco~Manuel Rangel~Pardo, Paolo Rosso, and Manuela Sanguinetti.
\newblock {S}em{E}val-2019 task 5: Multilingual detection of hate speech against immigrants and women in {T}witter.
\newblock In {\em Proceedings of the 13th International Workshop on Semantic Evaluation}, pages 54--63, Minneapolis, Minnesota, USA, 2019. Association for Computational Linguistics.

\bibitem[BCCEAN20]{tweeteval_all}
Francesco Barbieri, Jose Camacho-Collados, Luis Espinosa-Anke, and Leonardo Neves.
\newblock {TweetEval:Unified Benchmark and Comparative Evaluation for Tweet Classification}.
\newblock In {\em Proceedings of Findings of EMNLP}, 2020.

\bibitem[BDSMGN20]{allsides_dataset}
Ramy Baly, Giovanni Da~San~Martino, James Glass, and Preslav Nakov.
\newblock We can detect your bias: Predicting the political ideology of news articles.
\newblock In {\em Proceedings of the 2020 Conference on Empirical Methods in Natural Language Processing (EMNLP)}, EMNLP~'20, pages 4982--4991, 2020.

\bibitem[BEACC22]{tweetsentimentmultilingual}
Francesco Barbieri, Luis Espinosa~Anke, and Jose Camacho-Collados.
\newblock {XLM}-{T}: Multilingual language models in {T}witter for sentiment analysis and beyond.
\newblock In {\em Proceedings of the Thirteenth Language Resources and Evaluation Conference}, pages 258--266, Marseille, France, June 2022. European Language Resources Association.

\bibitem[BG24]{Battle2024TheUE}
Rick Battle and Teja Gollapudi.
\newblock The unreasonable effectiveness of eccentric automatic prompts.
\newblock {\em ArXiv}, abs/2402.10949, 2024.

\bibitem[BLHJ16]{barrett2016handbook}
L.F. Barrett, M.~Lewis, and J.M. Haviland-Jones.
\newblock {\em Handbook of Emotions, Fourth Edition}.
\newblock Psychology (The Guilford Press). Guilford Publications, 2016.

\bibitem[BMR{\etalchar{+}}20]{few_shot_learners}
Tom Brown, Benjamin Mann, Nick Ryder, Melanie Subbiah, Jared~D Kaplan, Prafulla Dhariwal, Arvind Neelakantan, Pranav Shyam, Girish Sastry, Amanda Askell, Sandhini Agarwal, Ariel Herbert-Voss, Gretchen Krueger, Tom Henighan, Rewon Child, Aditya Ramesh, Daniel Ziegler, Jeffrey Wu, Clemens Winter, Chris Hesse, Mark Chen, Eric Sigler, Mateusz Litwin, Scott Gray, Benjamin Chess, Jack Clark, Christopher Berner, Sam McCandlish, Alec Radford, Ilya Sutskever, and Dario Amodei.
\newblock Language models are few-shot learners.
\newblock In H.~Larochelle, M.~Ranzato, R.~Hadsell, M.F. Balcan, and H.~Lin, editors, {\em Advances in Neural Information Processing Systems}, volume~33, pages 1877--1901. Curran Associates, Inc., 2020.

\bibitem[CKD24]{RePrompt}
Weizhe Chen, Sven Koenig, and Bistra Dilkina.
\newblock Reprompt: Planning by automatic prompt engineering for large language models agents.
\newblock 06 2024.

\bibitem[DKG22]{DoanParty}
Tu~My Doan, Benjamin Kille, and Jon~Atle Gulla.
\newblock Using language models for classifying the party affiliation of political texts.
\newblock In Paolo Rosso, Valerio Basile, Raquel Mart{\'i}nez, Elisabeth M{\'e}tais, and Farid Meziane, editors, {\em Natural Language Processing and Information Systems}, pages 382--393, Cham, 2022. Springer International Publishing.

\bibitem[DMB23]{devatine:hal-04249724}
Nicolas Devatine, Philippe Muller, and Chlo{\'e} Braud.
\newblock {An Integrated Approach for Political Bias Prediction and Explanation Based on Discursive Structure}.
\newblock In {\em {Findings of the Association for Computational Linguistics (EACL 2023)}}, pages 11196--11211, Toronto, Canada, July 2023. {ACL: Association for Computational Linguistics}.

\bibitem[ea23]{Achiam2023GPT4TR}
OpenAI Josh~Achiam et~al.
\newblock Gpt-4 technical report.
\newblock 2023.

\bibitem[GAK23]{gilardi2023chatgpt}
Fabrizio Gilardi, Meysam Alizadeh, and Ma{\"e}l Kubli.
\newblock Chatgpt outperforms crowd workers for text-annotation tasks.
\newblock {\em Proceedings of the National Academy of Sciences}, 120(30):e2305016120, 2023.

\bibitem[Gil23]{jacobgildenblatconfidenceinterval}
Jacob Gildenblat.
\newblock A python library for confidence intervals.
\newblock \url{https://github.com/jacobgil/confidenceinterval}, 2023.

\bibitem[GL12]{howpartisanispress2012}
JOSHUA~S. GANS and ANDREW LEIGH.
\newblock How partisan is the press? multiple measures of media slant*.
\newblock {\em Economic Record}, 88(280):127--147, 2012.

\bibitem[HCvH24]{heseltine2024large}
Michael Heseltine and Bernhard Clemm~von Hohenberg.
\newblock Large language models as a substitute for human experts in annotating political text.
\newblock {\em Research \& Politics}, 11(1):20531680241236239, 2024.

\bibitem[HLG{\etalchar{+}}24]{he-etal-2024-annollm}
Xingwei He, Zhenghao Lin, Yeyun Gong, A-Long Jin, Hang Zhang, Chen Lin, Jian Jiao, Siu~Ming Yiu, Nan Duan, and Weizhu Chen.
\newblock {A}nno{LLM}: Making large language models to be better crowdsourced annotators.
\newblock In Yi~Yang, Aida Davani, Avi Sil, and Anoop Kumar, editors, {\em Proceedings of the 2024 Conference of the North American Chapter of the Association for Computational Linguistics: Human Language Technologies (Volume 6: Industry Track)}, pages 165--190, Mexico City, Mexico, June 2024. Association for Computational Linguistics.

\bibitem[HZD{\etalchar{+}}21]{Han2021PTRPT}
Xu~Han, Weilin Zhao, Ning Ding, Zhiyuan Liu, and Maosong Sun.
\newblock Ptr: Prompt tuning with rules for text classification.
\newblock {\em ArXiv}, abs/2105.11259, 2021.

\bibitem[JD16]{MLold1}
Anuja~P Jain and Padma Dandannavar.
\newblock Application of machine learning techniques to sentiment analysis.
\newblock In {\em 2016 2nd International Conference on Applied and Theoretical Computing and Communication Technology (iCATccT)}, pages 628--632, 2016.

\bibitem[JPZ{\etalchar{+}}24]{APEER}
Can Jin, Hongwu Peng, Shiyu Zhao, Wang Zhenting, Wujiang Xu, Ligong Han, Jiahui Zhao, Kai Zhong, Sanguthevar Rajasekaran, and Dimitris Metaxas.
\newblock Apeer: Automatic prompt engineering enhances large language model reranking.
\newblock 06 2024.

\bibitem[KGR{\etalchar{+}}22]{kojima2022large}
Takeshi Kojima, Shixiang~Shane Gu, Machel Reid, Yutaka Matsuo, and Yusuke Iwasawa.
\newblock Large language models are zero-shot reasoners.
\newblock {\em Advances in neural information processing systems}, 35:22199--22213, 2022.

\bibitem[LDC{\etalchar{+}}22]{Lampinen2022CanLM}
Andrew~Kyle Lampinen, Ishita Dasgupta, Stephanie C.~Y. Chan, Kory Matthewson, Michael~Henry Tessler, Antonia Creswell, James~L. McClelland, Jane~X. Wang, and Felix Hill.
\newblock Can language models learn from explanations in context?
\newblock {\em ArXiv}, abs/2204.02329, 2022.

\bibitem[LHM{\etalchar{+}}23]{ChatGPT}
Yiheng Liu, Tianle Han, Siyuan Ma, Jiayue Zhang, Yuanyuan Yang, Jiaming Tian, Hao He, Antong Li, Mengshen He, Zhengliang Liu, Zihao Wu, Lin Zhao, Dajiang Zhu, Xiang Li, Ning Qiang, Dingang Shen, Tianming Liu, and Bao Ge.
\newblock Summary of chatgpt-related research and perspective towards the future of large language models.
\newblock {\em Meta-Radiology}, 1(2):100017, 2023.

\bibitem[Li24]{Li2024comparative}
Jiyi Li.
\newblock A comparative study on annotation quality of crowdsourcing and llm via label aggregation.
\newblock pages 6525--6529, 04 2024.

\bibitem[lib]{libcon}
{Liberals vs Conservatives on Reddit}.
\newblock \url{https://www.kaggle.com/datasets/neelgajare/liberals-vs-conservatives-on-reddit-13000-posts}.
\newblock Accessed: 2024-06-16.

\bibitem[LJW{\etalchar{+}}22]{bias1}
Ruibo Liu, Chenyan Jia, Jason Wei, Guangxuan Xu, and Soroush Vosoughi.
\newblock Quantifying and alleviating political bias in language models.
\newblock {\em Artificial Intelligence}, 304:103654, 2022.

\bibitem[LLC{\etalchar{+}}24]{Political-LLM-multipleauthors}
Lincan Li, Jiaqi Li, Catherine Chen, Fred Gui, Hongjia Yang, Chenxiao Yu, Zhengguang Wang, Jianing Cai, Junlong Zhou, Bolin Shen, Alex Qian, Weixin Chen, Zhongkai Xue, Lichao Sun, Lifang He, Hanjie Chen, Kaize Ding, Zijian Du, Fangzhou Mu, and Yushun Dong.
\newblock Political-llm: Large language models in political science, 12 2024.

\bibitem[LPL{\etalchar{+}}22]{QianFromTraditionaltoDeepLearning}
Qian Li, Hao Peng, Jianxin Li, Congying Xia, Renyu Yang, Lichao Sun, Philip Yu, and Lifang He.
\newblock A survey on text classification: From traditional to deep learning.
\newblock {\em ACM Transactions on Intelligent Systems and Technology}, 13:1--41, 04 2022.

\bibitem[MBMSK18]{tweeteval_emotion}
Saif Mohammad, Felipe Bravo-Marquez, Mohammad Salameh, and Svetlana Kiritchenko.
\newblock Semeval-2018 task 1: Affect in tweets.
\newblock In {\em Proceedings of the 12th international workshop on semantic evaluation}, pages 1--17, 2018.

\bibitem[MPNR24]{bias2}
Fabio Motoki, Valdemar Pinho~Neto, and Victor Rodrigues.
\newblock More human than human: Measuring chatgpt political bias.
\newblock {\em Public Choice}, 198(1):3--23, 2024.

\bibitem[NR13]{MLold2}
M~S Neethu and R~Rajasree.
\newblock Sentiment analysis in twitter using machine learning techniques.
\newblock In {\em 2013 Fourth International Conference on Computing, Communications and Networking Technologies (ICCCNT)}, pages 1--5, 2013.

\bibitem[Ope23]{GPT4}
OpenAI.
\newblock Gpt-4 technical report.
\newblock 2023.

\bibitem[OSMC23]{OllionMindllmhype}
Etienne Ollion, Rubing Shen, Ana Macanovic, and Arnault Chatelain.
\newblock Chatgpt for text annotation? mind the hype!, 10 2023.

\bibitem[PP11]{pennacchiotti2011machine}
Marco Pennacchiotti and Ana-Maria Popescu.
\newblock A machine learning approach to twitter user classification.
\newblock In {\em Proceedings of the international AAAI conference on web and social media}, volume~5, pages 281--288, 2011.

\bibitem[pro]{prompt_engineering_guide}
Prompt engineering guide.
\newblock \url{https://www.promptingguide.ai/techniques/}.
\newblock Accessed: 01-06-2024.

\bibitem[RBOP24]{rasmussen2024super}
Stig Hebbelstrup~Rye Rasmussen, Alexander Bor, Mathias Osmundsen, and Michael~Bang Petersen.
\newblock ‘{S}uper-unsupervised’classification for labelling text: online political hostility as an illustration.
\newblock {\em British Journal of Political Science}, 54(1):179--200, 2024.

\bibitem[RFN17]{tweeteval_sentiment}
Sara Rosenthal, Noura Farra, and Preslav Nakov.
\newblock Semeval-2017 task 4: Sentiment analysis in twitter.
\newblock In {\em Proceedings of the 11th international workshop on semantic evaluation (SemEval-2017)}, pages 502--518, 2017.

\bibitem[SDZ23]{shum-etal-2023-automatic}
Kashun Shum, Shizhe Diao, and Tong Zhang.
\newblock Automatic prompt augmentation and selection with chain-of-thought from labeled data.
\newblock In Houda Bouamor, Juan Pino, and Kalika Bali, editors, {\em Findings of the Association for Computational Linguistics: EMNLP 2023}, pages 12113--12139, Singapore, December 2023. Association for Computational Linguistics.

\bibitem[SM10]{Strapparava2010}
Carlo Strapparava and Rada Mihalcea.
\newblock {\em Annotating and Identifying Emotions in Text}, pages 21--38.
\newblock Springer Berlin Heidelberg, Berlin, Heidelberg, 2010.

\bibitem[SSS{\etalchar{+}}24]{survey1}
Pranab Sahoo, Ayush Singh, Sriparna Saha, Vinija Jain, Samrat Mondal, and Aman Chadha.
\newblock A systematic survey of prompt engineering in large language models: Techniques and applications.
\newblock 02 2024.

\bibitem[T{\"o}r23]{tornberg2023chatgpt}
Petter T{\"o}rnberg.
\newblock Chatgpt-4 outperforms experts and crowd workers in annotating political twitter messages with zero-shot learning.
\newblock {\em arXiv preprint arXiv:2304.06588}, 2023.

\bibitem[TYKK22]{ConfidenceIntervals}
Kanae Takahashi, Kouji Yamamoto, Aya Kuchiba, and Tatsuki Koyama.
\newblock Confidence interval for micro-averaged f1 and macro-averaged f1 scores.
\newblock {\em Applied Intelligence}, 52, 03 2022.

\bibitem[VD24]{survey2}
Shubham Vatsal and Harsh Dubey.
\newblock A survey of prompt engineering methods in large language models for different nlp tasks.
\newblock 07 2024.

\bibitem[VSP{\etalchar{+}}17]{attention}
Ashish Vaswani, Noam Shazeer, Niki Parmar, Jakob Uszkoreit, Llion Jones, Aidan~N. Gomez, Lukasz Kaiser, and Illia Polosukhin.
\newblock Attention is all you need.
\newblock 2017.

\bibitem[WC17]{WilkersonComputerizedtextanalysis}
John Wilkerson and Andreu Casas.
\newblock Large-scale computerized text analysis in political science: Opportunities and challenges.
\newblock {\em Annual Review of Political Science}, 20:529--544, 05 2017.

\bibitem[WR23]{EvalAllYouNeed}
Maximilian Weber and Merle Reichardt.
\newblock Evaluation is all you need. prompting generative large language models for annotation tasks in the social sciences. a primer using open models.
\newblock 12 2023.

\bibitem[WWS{\etalchar{+}}22]{Wei2022ChainOT}
Jason Wei, Xuezhi Wang, Dale Schuurmans, Maarten Bosma, Ed~Huai hsin Chi, F.~Xia, Quoc Le, and Denny Zhou.
\newblock Chain of thought prompting elicits reasoning in large language models.
\newblock {\em ArXiv}, abs/2201.11903, 2022.

\bibitem[YCX{\etalchar{+}}23]{gpt35}
Junjie Ye, Xuanting Chen, Nuo Xu, Can Zu, Zekai Shao, Shichun Liu, Yuhan Cui, Zeyang Zhou, Chao Gong, Yang Shen, Jie Zhou, Siming Chen, Tao Gui, Qi~Zhang, and Xuanjing Huang.
\newblock A comprehensive capability analysis of gpt-3 and gpt-3.5 series models.
\newblock {\em ArXiv}, abs/2303.10420, 2023.

\bibitem[YWL{\etalchar{+}}24]{optimizer2}
Chengrun Yang, Xuezhi Wang, Yifeng Lu, Hanxiao Liu, Quoc~V. Le, Denny Zhou, and Xinyun Chen.
\newblock Large language models as optimizers, 2024.

\bibitem[ZLS{\etalchar{+}}24]{bias3}
Travis Zack, Eric Lehman, Mirac Suzgun, Jorge~A Rodriguez, Leo~Anthony Celi, Judy Gichoya, Dan Jurafsky, Peter Szolovits, David~W Bates, Raja-Elie~E Abdulnour, et~al.
\newblock Assessing the potential of gpt-4 to perpetuate racial and gender biases in health care: a model evaluation study.
\newblock {\em The Lancet Digital Health}, 6(1):e12--e22, 2024.

\bibitem[ZMH{\etalchar{+}}22]{optimizer1}
Yongchao Zhou, Andrei Muresanu, Ziwen Han, Keiran Paster, Silviu Pitis, Harris Chan, and Jimmy Ba.
\newblock Large language models are human-level prompt engineers, 11 2022.

\bibitem[ZMN{\etalchar{+}}19]{tweetevaloffensive}
Marcos Zampieri, Shervin Malmasi, Preslav Nakov, Sara Rosenthal, Noura Farra, and Ritesh Kumar.
\newblock Semeval-2019 task 6: Identifying and categorizing offensive language in social media (offenseval).
\newblock In {\em Proceedings of the 13th International Workshop on Semantic Evaluation}, pages 75--86, 2019.

\end{thebibliography}

\newpage

 \appendix
 
\section{Results of the main experiments as numerical tables}
We report in Tables \ref{table:accuracies} and \ref{table:accuracies_train_vs_test} the same results as in Figures \ref{fig:main_results} and \ref{fig:test_vs_train}.

\renewcommand{\arraystretch}{1.2}
\begin{table}[h!]
	\centering
	\begin{tabular}{|c|c|c|c|c|}
		\multicolumn{5}{c}{\textbf{Micro averaged F1-scores}} \\
		\hline
		                                                     & TE-hate                      & TE-emotion                   & TE-sent                      & TE-off                   \\
		\hhline{|=|=|=|=|=|} Simple     & $\textit{57.0 [55.2, 58.8]}$ & $78.7 [76.6, 80.9]$          & $\textit{60.5 [59.5, 61.4]}$ & $\textit{71.4 [68.4, 74.4]}$      \\
		\hline
		Explanations                    & $62.7 [60.9, 64.4]$          & $79.5 [77.4, 81.6]$          & $68.3 [67.4, 69.2]$          & $80.1 [77.4, 82.8]$      \\
		\hline
		Examples                        & $61.3 [59.6, 63.1]$          & $\textit{76.9 [74.7, 79.1]}$ & $\bf{70.6 [69.7, 71.5]}$     & $72.9 [69.9, 75.9]$      \\
		\hline
		Roleplay                        & $64.7 [63.0, 66.5]$          & $79.7 [77.6, 81.8]$          & $68.3 [67.4, 69.2]$          & $\bf{80.6 [77.9, 83.2]}$ \\
		\hline
		CoT                             & $\bf{65.2 [63.4, 66.9]}$          & $79.0 [76.9, 81.1]$          & $63.3 [62.3, 64.2]$          & $72.9 [69.9, 75.9]$      \\
		\hline
		APO                             & $62.5 [60.6, 64.4]$     & $\bf{79.9 [77.5, 82.4]}$          & $70.0 [69.1, 70.9]$          & $75.0 [71.0, 79.0]$                    \\
		\hline
        \\
        \hhline{|-|-|-|-|}
                                                             & TML-sent                     & AS-pol                       & LibCon                       \\
        \hhline{|=|=|=|=|} Simple                            & $67.1 [66.0, 68.2]$          & $52.8 [51.8, 53.8]$          & $73.3 [72.4, 74.2]$          \\
        \hhline{|-|-|-|-|}
        Explanations                                         & $68.4 [67.4, 69.2]$          & $47.9 [46.9, 48.8]$          & $72.3 [71.4, 73.2]$          \\
        \hhline{|-|-|-|-|}
        Examples                                             & $66.0 [64.9, 67.1]$          & $52.3 [51.3, 53.3]$          & $73.6 [72.7, 74.4]$          \\
        \hhline{|-|-|-|-|}
        Roleplay                                             & $\textit{65.5 [64.4, 66.7]}$ & $\textit{47.1 [46.1, 48.1]}$ & $\textit{70.7 [69.8, 71.5]}$ \\
        \hhline{|-|-|-|-|}
        CoT                                                  & $67.7 [66.6, 68.8]$          & $47.9 [47.0, 48.9]$          & $73.9 [73.1, 74.8]$          \\
        \hhline{|-|-|-|-|}
        APO                                                  & $\bf{69.2 [68.1, 70.3]}$          & $\bf{55.4 [54.4, 56.4]}$     & $\bf{74.4 [73.5, 75.2]}$                  \\
        \hhline{|-|-|-|-|}
	\end{tabular}
	\vspace{5pt}
	\caption{ Micro-averaged $F_1$ scores  (in $\%$) of the hand-crafted prompts and of the best prompt
	obtained using automatic prompt optimization (APO) on each of the datasets and
	tasks described in Subsection \ref{subsec:datasets}, along with the lower bound and upper bound of the $95\%$ confidence interval of each measure (in brackets). The highest score achieved
	on a task is in bold, the lowest score is in italics. }
	\label{table:accuracies}
\end{table}

\renewcommand{\arraystretch}{1.2}
\begin{table}[h!]
	\centering
	\begin{tabular}{|c|c|c|c|c|}
		\multicolumn{5}{c}{\textbf{Train set vs test set}} \\
		\hline
		                                & TE-hate, train           & TE-hate, test                & TE-emotion, train                      & TE-emotion, test                   \\
		\hhline{|=|=|=|=|=|} Simple     & $76.3 [75.4, 77.1]$      & $57.0 [55.2, 58.8]$          & $ 76.9 [75.4, 78.3] $ &   $ 78.7 [76.6, 80.9]$      \\
		\hline
		Explanations                    & $ 77.6 [76.7, 78.4] $    & $62.7 [60.9, 64.4]$          & $75.9 [74.4, 77.4]$          & $79.5 [77.4, 81.6]$      \\
		\hline
		Examples                        &  $78.1 [77.2, 79.0]$     & $61.3 [59.6, 63.1]$          & $74.5 [73.0, 76.0]$     & $76.9 [74.7, 79.1]$      \\
		\hline
		Roleplay                        & $77.7 [76.8, 78.6]$      & $64.7 [63.9, 66.5]$          & $77.2 [75.8, 78.7]$          & $ 79.7 [77.6, 81.8]$ \\
		\hline
		CoT                             & $78.8 [78.0, 79.7]$      & $65.2 [63.4, 66.9]$          & $76.3 [74.9, 77.8]$          & $79.0 [76.9, 81.1]$      \\
		\hline
	\end{tabular}
	\vspace{5pt}
	\caption{ Micro-averaged $F_1$ scores  (in $\%$) of the hand-crafted prompts and of the best prompt
	obtained using automatic prompt optimization (APO) on each of the datasets and
	tasks described in Subsection \ref{subsec:datasets}, along with the lower bound and upper bound of the $95\%$ confidence interval of each measure (in brackets). The highest score achieved
	on a task is in bold, the lowest score is in italics. }
	\label{table:accuracies_train_vs_test}
\end{table}

\newpage

\section{Optimized prompts}
For each task, we report the best prompt generated by the prompt optimization process:

\textbf{TE-hate}
``Check for hate speech in the following message to determine if it is hateful, then classify it as either "hateful" or "non-hateful".''

\textbf{TE-emotion}
``Identify the emotion displayed in the following message as joy, anger, sadness or optimism.''

\textbf{TE-sent}
``Label the emotion in the given message as positive, negative, or neutral.''

\textbf{TE-off}
``Determine if the following message is offensive or non-offensive and provide the corresponding label.''

\textbf{TML-sent}
``Categorize the sentiment in the following message as positive, negative, or neutral.''

\textbf{AS-pol}
``Identify if the text below belongs to the left, center, or right categories.''

\textbf{LibCon}
``Identify the text as either liberal or conservative.''

\newpage

\section{Brief overview of our browser-based service} \label{app:online_app}
We summarily describe our automatic text labelling service
\url{https://prompt-ultra.github.io/}. Further explanations are given on the website itself.

\begin{figure}[h]
    \centering
    \includegraphics[width=0.8\textwidth]{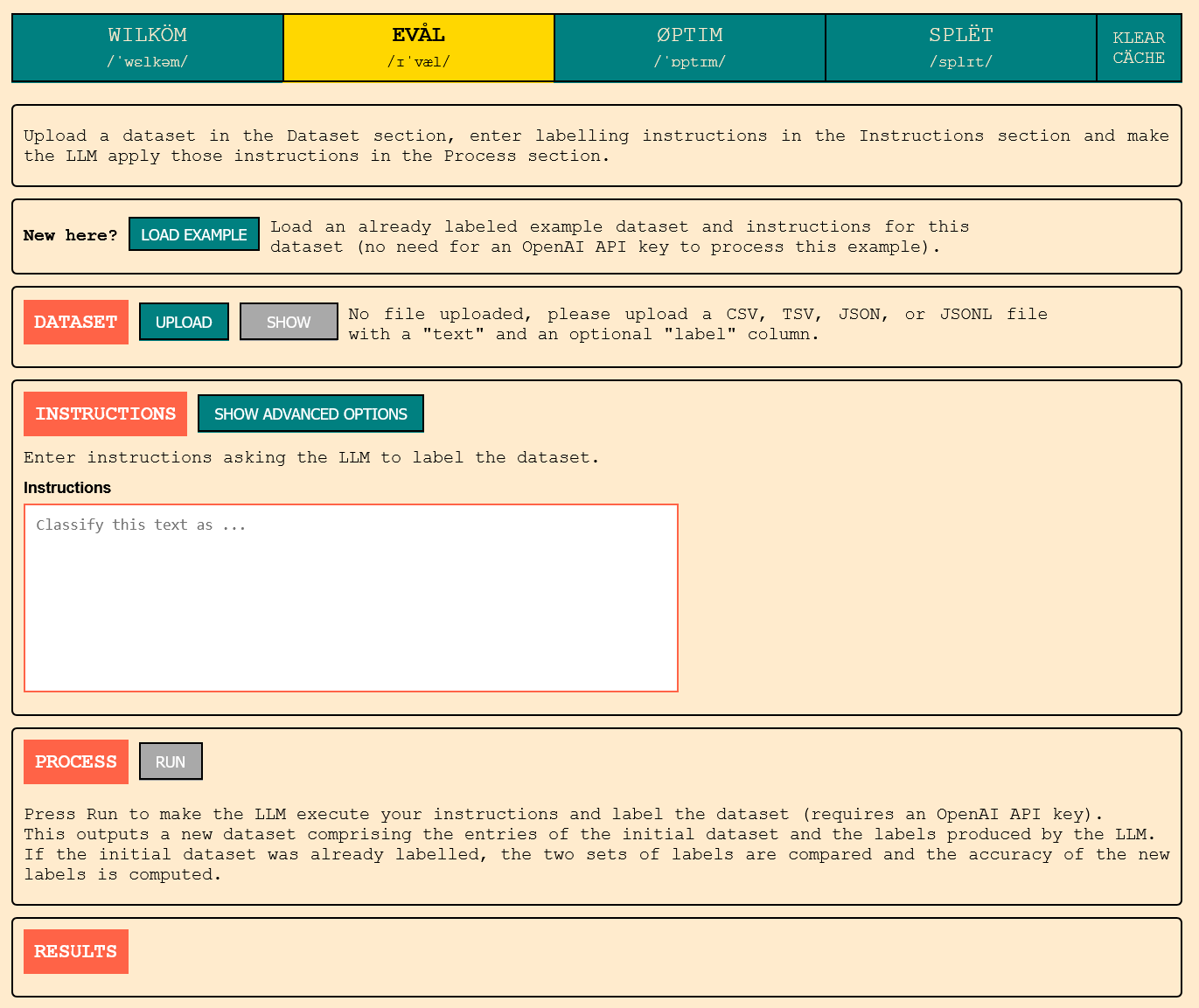}
    \caption{The \textbf{EV\r{A}L} tab.}
    \label{fig:Eval}
\end{figure}

In the \textbf{EV\r{A}L} tab (see Figure \ref{fig:Eval}), you upload a dataset, which can be labelled or unlabelled,  using the topmost button.
You then enter a prompt and press the \textit{Run} button. At that point, a pop-up window requires you to input your ChatGPT access key. Once you have done so, the prompt is applied to each entry of the dataset, and the resulting labelled dataset is output. If the original dataset was labelled, its labels and the predicted labels are compared and the resulting accuracy is computed.

\begin{figure}[h]
    \centering
    \includegraphics[width=0.7\textwidth]{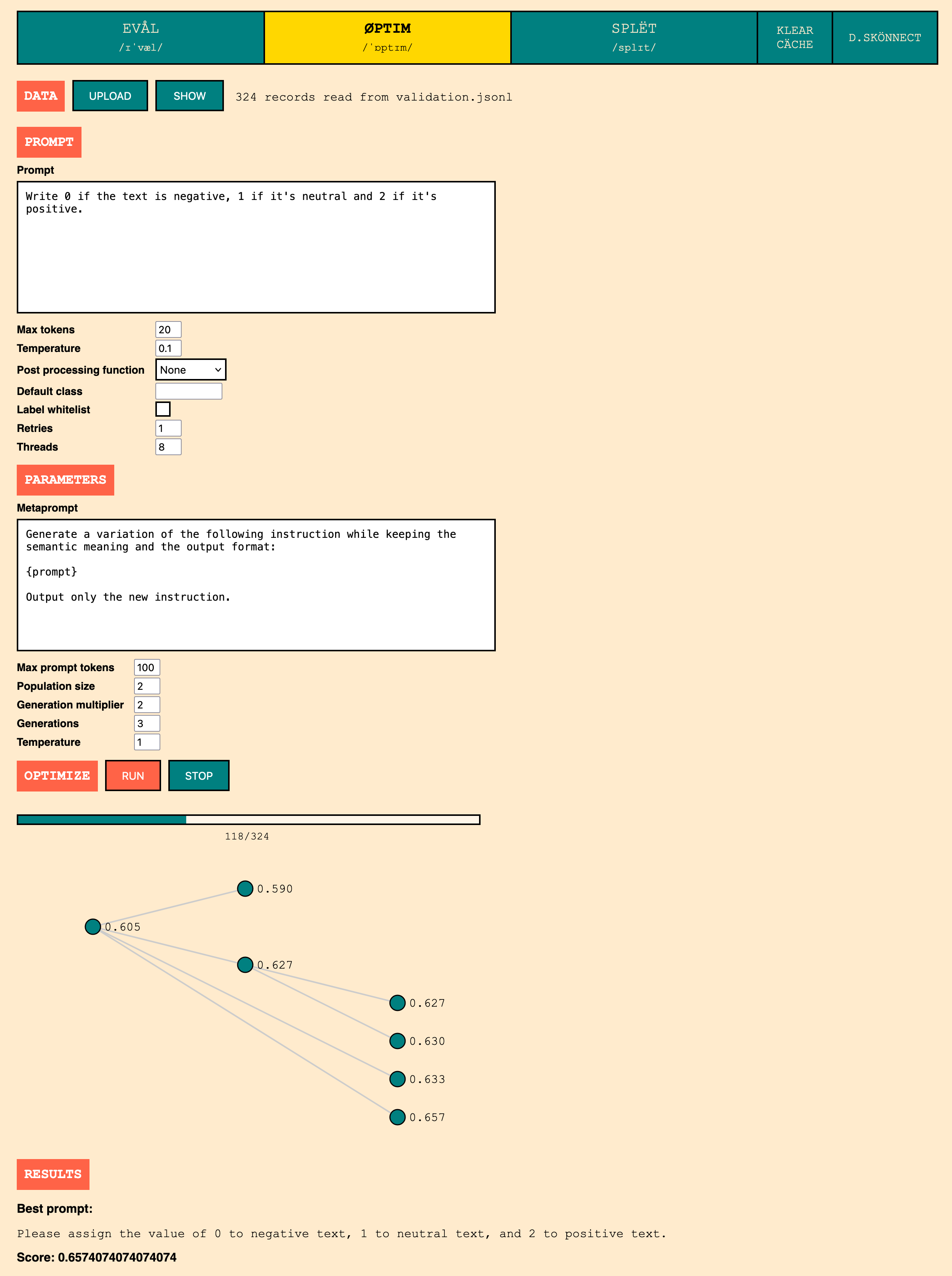}
    \caption{The \textbf{\O PTIM} tab.}
    \label{fig:Optim}
\end{figure}

Prompt optimization is performed in the \textbf{\O PTIM} tab (see Figure \ref{fig:Optim}). You first upload a labelled dataset using the topmost button, and input a starting prompt. A macroprompt is then applied to generate successive generations of prompts, with the best prompts of each generation surviving to the next generation. A graph illustrates which prompt descends from which prompt of the previous generation.
After a specified number of generations, the prompt whose performance was the best on the labelled dataset is output, as well as its accuracy score.

\begin{figure}[h]
    \centering
    \includegraphics[width=0.7\textwidth]{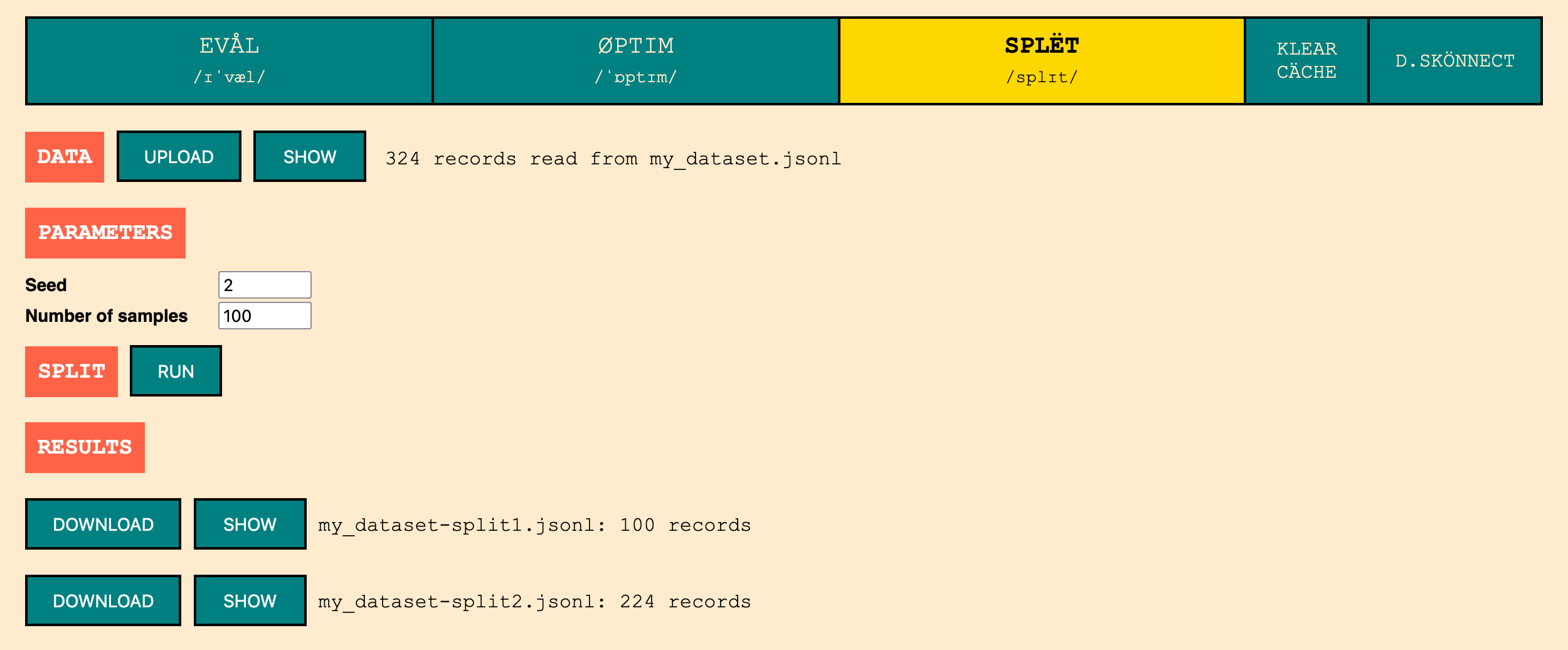}
   \caption{The \textbf{SPLËT} tab.}
    \label{fig:Split}
\end{figure}

The \textbf{SPLËT} tab (see Figure \ref{fig:Split}) simply provides a convenient way to split a dataset in two.

Finally, the \textbf{KLEAR CÄCHE} button allows you to clear the LLM's cache, and \textbf{D.SKÖNNECT} to erase the ChatGPT access key that you had entered.

\end{document}